\documentclass[10pt,twocolumn,letterpaper]{article}

\usepackage{cvpr}              % To produce the CAMERA-READY version
\definecolor{cvprblue}{rgb}{0.21,0.49,0.74}
\usepackage[pagebackref,breaklinks,colorlinks,allcolors=cvprblue]{hyperref}

\usepackage{booktabs,tabularx,makecell,subcaption,siunitx,xcolor}
\newcolumntype{L}{>{\raggedright\arraybackslash}X} % 左对齐可换行列
\usepackage{pifont}
\usepackage{multirow}
\usepackage[ruled,vlined]{algorithm2e}

\def\paperID{8498} % *** Enter the Paper ID here
\def\confName{CVPR}
\def\confYear{2026}

\title{TaCo: Capturing Spatio-Temporal Semantic Consistency in Remote Sensing Change Detection
}

\author{Han Guo, Chenyang Liu, Haotian Zhang, Bowen Chen, Zhengxia Zou,  Zhenwei Shi\\
Beihang University\\
}

\begin{document}
\maketitle
\begin{abstract}
Remote sensing change detection (RSCD) aims to identify surface changes across bi-temporal satellite images. Most previous methods rely solely on mask supervision, which effectively guides spatial localization but provides limited constraints on the temporal semantic transitions. Consequently, they often produce spatially coherent predictions while still suffering from unresolved semantic inconsistencies. To address this limitation, we propose TaCo, a spatio-temporal semantic consistent network, which enriches the existing mask-supervised framework with a spatio-temporal semantic joint constraint. TaCo conceptualizes change as a semantic transition between bi-temporal states, in which one temporal feature representation can be derived from the other via dedicated transition features. To realize this, we introduce a Text-guided Transition Generator that integrates textual semantics with bi-temporal visual features to construct the cross-temporal transition features. In addition, we propose a spatio-temporal semantic joint constraint consisting of bi-temporal reconstruct constraints and a transition constraint: the former enforces alignment between reconstructed and original features, while the latter enhances discrimination for changes. This design can yield substantial performance gains without introducing any additional computational overhead during inference. Extensive experiments on six public datasets, spanning both binary and semantic change detection tasks, demonstrate that TaCo consistently achieves SOTA performance.
\end{abstract}

% As a result, such methods can produce spatially well-structured outputs, but semantic errors are difficult to correct effectively.     
\section{Introduction}
\label{sec:intro}

Remote sensing change detection (RSCD) aims to identify land-cover changes by analyzing satellite or aerial images of the same region acquired at different times.
It has been widely applied in urban planning \cite{urban2, urban3}, environmental monitoring \cite{envir}, precision agriculture \cite{liu2022mscanet, crop2}, resource management \cite{resource} and disaster assessment \cite{gupta2019xbd, weber2020damformer}.

% \begin{figure}
% \centering
% \includegraphics[width=3.6in]{fig/intro_b.png}
% \caption{Visualization results of the differential feature maps on the SECOND test set.}
% \label{feature_vis}
% \end{figure}

\begin{figure}[t]
    \centering

    % ---------- 子图 (b) ----------
    \begin{subfigure}{0.99\linewidth}
        \centering
        \includegraphics[width=\linewidth]{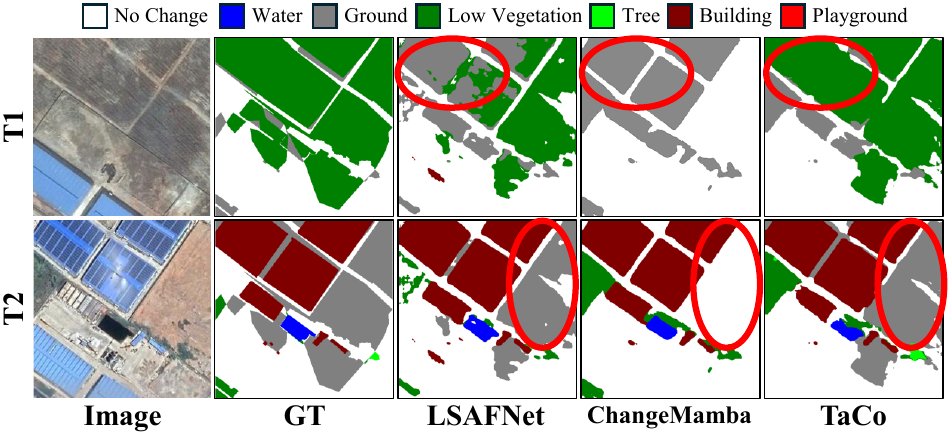}
        \caption{Semantic confusion in existing methods.}
        \label{fig:a}
    \end{subfigure}
    \vspace{2mm} % 两个子图之间的间距

    % ---------- 子图 (a) ----------
    \begin{subfigure}{0.95\linewidth}
        \centering
        \includegraphics[width=\linewidth]{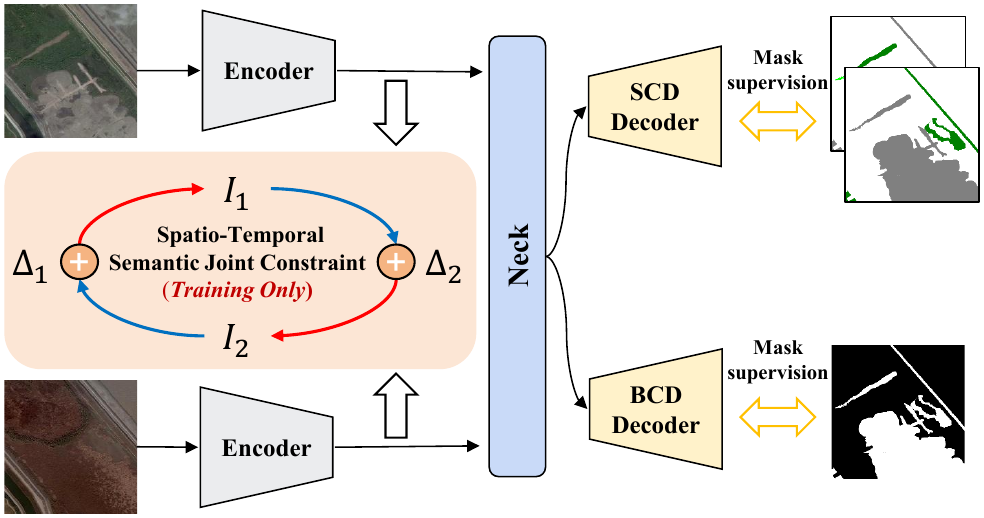}
        \caption{Our spatio-temporal semantic consistent network.}
        \label{fig:b}
    \end{subfigure}
    \vspace{-3mm}
    \caption{(a) Existing methods suffer from semantic misclassification and pseudo-changes. (b) Our method introduces a spatio-temporal semantic joint constraint during training.}
    \label{fig:vertical_two}
    \vspace{-6mm}
\end{figure}

Existing RSCD methods typically adopt a siamese architecture, where weight-shared encoders extract bi-temporal features in a shared semantic space for subsequent comparison or fusion to predict changes.
Recent studies \cite{zhu2025change3d,bandara2022changeformer, chen2024ttp, zhang2024bifa, fang2021snunet, wang2025ckcd, yuan2022landsat, chen2024changemamba, zhao2024rsmamba, ding2022bisrnet} have sought to improve this architecture from two main perspectives.
(i) Employing better backbones to improve generalization and high-level semantic representation.
(ii) Designing more ingenious interaction modules to better capture temporal differences between bi-temporal features.

Although the field of change detection has witnessed rapid progress, most methods are designed to be trained with only mask supervision, which remains fundamentally spatial as it enforces correspondence with discrete pixel-level label indices.
This supervision effectively guides the network to localize changed and unchanged regions. However, it 
provides weak constraints on the underlying temporal semantic transitions, where one land-cover state transitions into another. Consequently, these methods often produce spatially coherent predictions while still suffering from unresolved semantic inconsistencies.
% As a result, existing methods can produce spatially well-structured outputs, but semantic errors are difficult to correct effectively.

To address this limitation, we propose TaCo, a spatio-temporal semantic consistent network, which enriches the existing mask-supervised framework with a spatio-temporal semantic joint constraint.
TaCo provides a new perspective by treating change as a semantic transition process, in which one temporal feature state can be conceptually reconstructed from the other under the guidance of textual semantics.
During training, textual priors derived from dataset's category names (e.g., building, ground, water) act as semantic anchors, guiding the temporal transition between bi-temporal features.
Importantly, TaCo requires no additional annotations or textual input during inference, preserving both computational efficiency.

To realize the proposed semantic transition process, we design a Text-guided Transition Generator that integrates textual semantics with bi-temporal visual features to construct cross-temporal transition features.
In addition, we propose a spatio-temporal semantic joint constraint composed of bi-temporal reconstruction constraints and a transition constraint.
Bi-temporal reconstruction constraints align reconstructed features with their original counterparts, modeling temporal coherence across images.
The transition constraint improves discriminability in changed regions by pushing apart their representations, while simultaneously preserving semantic consistency in unchanged regions by pulling their features closer.
Collectively, these components allow TaCo to jointly learn spatial localization and temporal semantic consistency within a unified training framework.

Our main contributions are summarized as follows:
\begin{itemize}[leftmargin=1.2em, labelsep=0.5em]
\item We present TaCo, offering a new perspective for mask-supervised RSCD by formulating change as a semantic transition between temporal states.

\item We propose a Text-guided Transition Generator to construct transition features and a spatio-temporal semantic joint constraint that improves performance without introducing any inference-time overhead.
% We propose a Text-guided Transition Generator to construct transition features, and a spatio-temporal semantic joint constraint to improve the model’s performance without introducing any additional computational overhead during inference.
\item Experiments on six datasets demonstrate that TaCo achieves SOTA performance. Furthermore, we conduct an in-depth analysis of the proposed consistency modeling, offering insights for future research.
\end{itemize}

\section{Related Work}

% \textbf{Remote Sensing Change Detection.} Existing RSCD methods typically adopt a siamese architecture, where weight-shared encoders extract bi-temporal features in a shared semantic space for subsequent comparison or fusion to predict changes. Recent studies have sought to improve this architecture from two main perspectives.
Previous RSCD studies mainly focused on improving performance from two main perspectives:

\textbf{(i) Better Backbones.} Early RSCD frameworks were primarily built upon convolutional networks, which provide strong local modeling capability and accurate boundary delineation \cite{zhang2020ifnet, fang2021snunet, daudt2019hrscd, ding2022bisrnet, liu2022mscanet, li2024defo}.
However, the limited receptive field and weak global reasoning of CNNs hinder their ability to capture cross-temporal dependencies and high-level semantic relations, especially when changes occur gradually or involve large-scale contextual shifts.
To overcome these limitations, Transformer-based architectures \cite{vaswani201transformer} have been introduced to exploit long-range dependencies and global context modeling, achieving better generalization across diverse scenes \cite{bandara2022changeformer, zhang2024bifa, wang2024cdsc, yuan2022landsat, zhang2022swinsunet}.
Yet, their quadratic computational complexity with respect to image resolution restricts their scalability to large remote sensing scenes.
Recent studies have therefore explored more efficient sequence-modeling paradigms, such as state-space models like Mamba \cite{gu2024mamba}, which maintain global interaction while reducing computation to linear complexity \cite{chen2024changemamba, zhao2024rsmamba, zhang2025cdmamba, zhang2025foba}.
Concurrently, foundation models such as CLIP \cite{radford2021clip}, SAM \cite{kirillov2023sam}, and Dinov3 \cite{simeoni2025dinov3} have been introduced to transfer high-level semantic priors from natural images to remote sensing data, thereby improving the model's ability to recognize complex land-cover categories and subtle changes \cite{mei2024scd-sam, zhang2025re-sam, dong2024changeclip, dong2025peftcd, chen2024ttp}.

\begin{figure*}[t]
\centering
\includegraphics[width=1\linewidth]{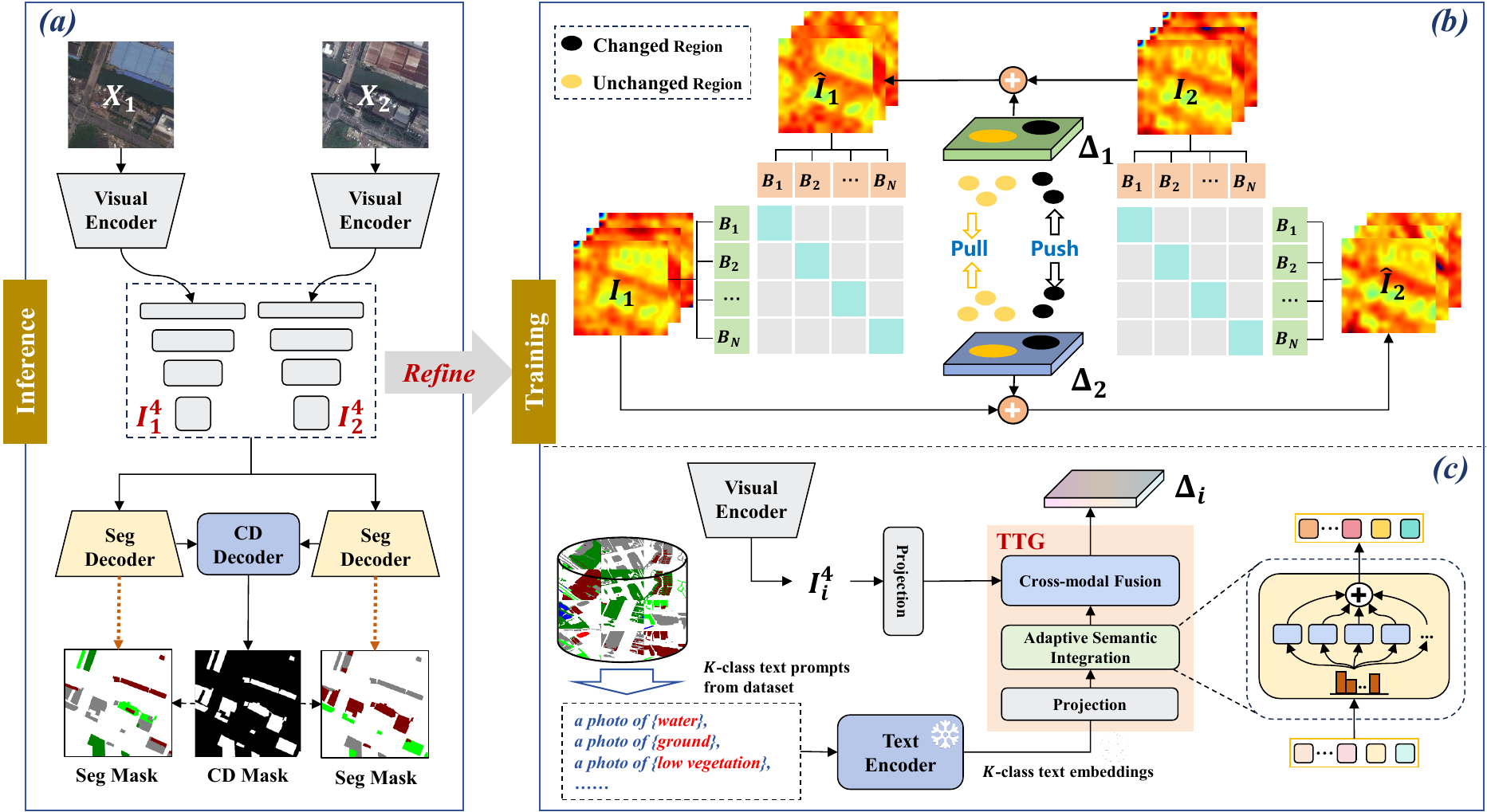}
\vspace{-6mm}
\caption{Overview of the proposed TaCo.
(a) Structure and inference pipeline based on the siamese encoder–decoder.
(b) Spatio-temporal semantic joint constraint on high-level features via reconstruction and transition losses.
(c) Text-guided Transition Generator that fuses class-level text embeddings with stage-4 visual tokens to construct transition features $\Delta_i$.}
\label{overview}
\vspace{-6mm}
\end{figure*}

\textbf{(ii) Ingenious Interaction Modules.}
Beyond employing better backbones, a substantial line of research focuses on enhancing bi-temporal interaction.
At the feature level, many methods introduce cross-temporal attention or dual-branch fusion blocks to strengthen correspondence between phases and refine object boundaries, thereby alleviating local spatial ambiguity \cite{li2022darnet,song2022acabfnet,zhang2020ifnet,fang2021snunet}.
Beyond generic attention-based fusion, some works explicitly estimate inter-temporal offsets before feature aggregation \cite{zhang2024bifa}. 
% For example, alignment modules such as BiFA compensate for spatial misregistration by learning pixel-wise or region-wise shifts between bi-temporal features \cite{zhang2024bifa}. 
In parallel, late-stage fusion strategies refine predictions with high-level semantics to suppress appearance-induced noise \cite{li2024defo,zhou2024lsafnet}.
At the semantic level, interaction is modeled in terms of category-aware reasoning rather than raw feature matching.
Multi-task networks jointly predict per-phase semantic maps and change masks, enforcing mutual constraints between the two tasks and enhancing category-level temporal understanding \cite{daudt2019hrscd,ding2022bisrnet,li2024defo,wang2024cdsc,ding2024scannet}.
More recently, vision-language paradigms have been introduced to further enrich temporal interaction. 
In particular, text-conditioned representations or change-aware knowledge are leveraged to guide bi-temporal fusion and enhance semantic reasoning under complex appearance variations \cite{dong2024changeclip,wang2025ckcd}.

Despite notable progress, most RSCD methods remain constrained by mask-level supervision, focusing on spatial localization rather than temporal semantic evolution.
We provide a new perspective by viewing change as a semantic transition and introducing spatio-temporal semantic modeling to explicitly capture such transitions across time.

\section{Methodology}

In this section, we will introduce the proposed TaCo, a spatio-temporal semantic consistent network, for RSCD. TaCo extends the existing mask-supervised framework by introducing a spatio-temporal semantic joint constraint. We leverage text prompts to guide temporal feature evolution, which enables the model to capture how land-cover semantics transition across time rather than merely focusing on spatial mask distribution.
We next describe the proposed method from four perspectives:  
(i) structure of TaCo that adopts a weight-shared encoder and a three-branch decoding architecture (Sec.~\ref{sec:structure});  
(ii) a spatio-temporal semantic joint constraint that enables model to jointly learn spatial localization and temporal semantic consistency (Sec.~\ref{sec:stsm});  
(iii) a Text-guided Transition Generator that integrates textual semantics with bi-temporal visual features to construct cross-temporal transition features (Sec.~\ref{sec:ttg}); and  
(iv) the training objective and loss functions (Sec.~\ref{sec:loss}).

\subsection{Structure of TaCo}\label{sec:structure}
As shown in Fig.~\ref{overview}, our model follows a mask-supervised siamese encoder–decoder architecture designed for both binary and semantic change detection. 
Given a pair of bi-temporal images $X_1$ and $X_2$, a weight-shared SegFormer~\cite{xie2021segformer} encoder extracts multi-level features for each temporal phase. 
The encoded features are then decoded through two auxiliary semantic decoders and a central change detection decoder. 
The semantic decoders refine per-phase semantic features that serve as feature guidance for the CD decoder and further generate semantic segmentation maps for SCD. 
The CD decoder integrates those refined features to produce the final change maps. 

Built upon this structure, we further propose a spatio-temporal semantic consistency training strategy that effectively improves the model's performance without introducing additional computational overhead during inference.

\subsection{Spatio-Temporal Semantic Modeling}\label{sec:stsm}
As shown in Fig.~\ref{overview}, our method views change as a temporal semantic transition rather than a purely spatial difference. The core idea is that one temporal state can be conceptually derived from the other through the learned transition features, enabling the model to capture cross-temporal semantic relationships. This modeling approach consists of two tightly coupled components: a spatio-temporal semantic joint constraint (panel~b) that leverages transition features to enforce temporal semantic consistency between temporal features, and a Text-guided Transition Generator (panel~c) that constructs token-wise transition features under textual semantics for consistent spatio-temporal modeling.
Importantly, the proposed approach is employed only during training, keeping the inference pipeline entirely unchanged and computationally lightweight.

Let $I_1^4$ and $I_2^4$ denote the high-level features of $X_1$ and $X_2$ extracted by the weight-shared visual encoder (stage-4 tokens in practice). 
Based on the change category annotations provided by the dataset, we construct text prompts (e.g., ``\textit{a photo of building}", ``\textit{a photo of water}"), which are encoded by the Git-RSCLIP text encoder~\cite{liu2025text2earth} to obtain class embeddings $T_{\text{class}}$. 
The Text-guided Transition Generator (TTG) (Sec.~\ref{sec:ttg}) then projects $T_{\text{class}}$ into a unified semantic space and integrates it with $I_i^4$ ($i \in \{1,2\}$) through Adaptive Semantic Integration and Cross-modal Fusion modules, producing cross-temporal transition features $\Delta_i$ for spatio-temporal semantic modeling.
% Two symmetric mappings are then performed to reconstruct each temporal feature from the other together with its transition features:
With these transition features, we further define two symmetric mappings that reconstruct each temporal feature from the other along with its corresponding transition features:
\begin{equation}
\label{equ_1}
I_1^4 + \Delta_2 = \hat{I}_2^4 \Rightarrow I_2^4
\vspace{-1mm}
\end{equation}
\begin{equation}
\label{equ_2}
I_2^4 + \Delta_1 = \hat{I}_1^4 \Rightarrow I_1^4
\vspace{-1mm}
\end{equation}
thereby enabling the model to reason about semantic correspondences in both temporal directions.

To enforce the spatio-temporal joint constraint, we propose two complementary token-level contrastive objectives. 
Bi-temporal reconstruction constraints encourage mutual alignment between reconstructed and original features, maintaining temporal consistency and preserving the semantic integrity of both temporal states.
The transition constraint aims to pull the representations of unchanged regions closer across time while pushing those of changed regions farther apart, thereby enhancing temporal discriminability. 
Through the spatio-temporal semantic modeling, TaCo preserves semantic consistency in unchanged regions and sharpens category transitions in changed regions. This effectively improves temporal reasoning without introducing additional computational overhead during inference.

\subsection{Text-guided Transition Generator}\label{sec:ttg}
TTG is proposed to obtain token-level transition features $\Delta_i$ that guide spatio-temporal semantic modeling. 
Rather than relying solely on visual cues, TTG leverages textual priors as category-level anchors, which provide global semantics across scenes and temporal phases. 
This module facilitates the distinction of visually similar changes (e.g., vegetation $\to$ ground) under class-aware guidance.
% , and it is only used during training.

\textbf{(i) Adaptive Semantic Integration.}
We employ the text encoder from Git-RSCLIP, a foundation model pretrained on global-scale remote sensing image-text pairs, to obtain class-level textual embeddings $T_{\text{class}}\!\in\!\mathbb{R}^{B\times K\times D_t}$, where $K$ denotes the number of dataset's label categories and $D_t$ represents the embedding dimension.
Since text prompts encode high-level categorical semantics, whereas visual tokens primarily capture local spatial patterns, directly fusing frozen textual embeddings with visual tokens often results in suboptimal cross-modal alignment due to the inherent modality gap.
To mitigate this issue, we introduce a lightweight Soft Mixture-of-Experts (SoftMoE) that adaptively projects $T_{\text{class}}$ into a unified semantic space. 

Specifically, let $M$ be the number of experts. We first compute the soft expert weights as
\vspace{-1mm}
\begin{equation}
\alpha \;=\; \mathrm{Softmax}\!\big(\mathrm{LN}(T_{\text{class}})\,W_\alpha\big)
\;\in\; \mathbb{R}^{B\times K\times M}
\vspace{-1mm}
\end{equation}
where $W_\alpha\!\in\!\mathbb{R}^{D_t\times M}$. 
% And $E_m(\cdot)$ projects textual features from $\mathbb{R}^{D_t}$ into the shared semantic space $\mathbb{R}^{D}$ with learnable parameters $W_m$.
And each expert $E_m(\cdot)$ is a learned nonlinear projection in the shared semantic space $\mathbb{R}^D$. 
The final textual embedding is obtained via soft aggregation
\vspace{-3mm}
\begin{equation}
Z \;=\; \sum_{m=1}^{M} \alpha_{:,:,m} \odot E_m(T_{\text{class}})
\;\in\; \mathbb{R}^{B\times K\times D}
\vspace{-2mm}
\end{equation}
where $\odot$ denotes element-wise scaling with broadcasting over the channel dimension. 
This multi-expert mechanism dynamically activates semantic subspaces and aligns text-derived semantics to the unified space.
% , thereby improving the cross-modal compatibility of textual and visual representations.

\textbf{(ii) Cross-modal Fusion.}
Let $I_i^4\!\in\!\mathbb{R}^{B\times N\times D_v}$ be the stage-4 visual tokens of the $i$-th backbone ($i\!\in\!\{1,2\}$), where $N$ is the number of tokens and $D_v$ is the visual channel dimension. 
We first project $I_i^4$ into the unified dimension $D$ with a linear projection $P_v\!\in\!\mathbb{R}^{D_v\times D}$, yielding $V^{(0)} \!=\! I_i^4 P_v \in \mathbb{R}^{B\times N\times D}$. 
We then feed the above textual embedding $Z$ as external semantic guidance into a transformer decoder to modulate $V^{(0)}$ and produce transition features $\Delta_i$. 
The decoder consists of $L$ stacked layers. Each layer $\ell = 1,\ldots,L$ performs self-attention on visual tokens followed by cross-attention between visual and textual representations
\vspace{-2mm}
\begin{equation}
V^{(\ell)} = \mathrm{SelfAttn}\!\left(V^{(\ell-1)}\right)
+ \mathrm{CrossAttn}\!\left(V^{(\ell-1)}, Z\right)
\vspace{-2mm}
\end{equation}
where $V^{(\ell-1)}$ serves as queries, and $Z$ provides both keys and values for cross-attention. The final output $V^{(L)}$ is linearly projected back to the original visual dimension to obtain transition features $\Delta_i$.

Overall, TTG establishes a concise and effective text-to-vision semantic bridge: Adaptive Semantic Integration (ASI) calibrates textual semantics into a unified space, and Cross-modal Fusion injects these calibrated semantics into bi-temporal tokens to produce discriminative transition features $\Delta_i$. 
These features are subsequently used in the spatio-temporal semantic modeling stage to strengthen cross-temporal semantic consistency.

\subsection{Loss Function}\label{sec:loss}

We optimize TaCo with a spatial mask-supervised loss and two token-level regularization terms: bi-temporal reconstruction contrast on $(I_i^4,\hat{I}_i^4)$ and a transition constraint on $(\Delta_1,\Delta_2)$.  
In this way, the network retains strong spatial supervision from masks while explicitly mining spatio-temporal semantic consistency in the feature space.

\textbf{(i) Spatial Mask-Supervised Loss.}
Following prior work on BCD and SCD~\cite{chen2021bit,ding2022bisrnet}, we adopt a spatial mask-supervised objective
\begin{equation}
\mathcal{L}_{\text{cd}} =
\begin{cases}
\mathcal{L}_{\text{change}}, & \text{for BCD} \\[4pt]
\mathcal{L}_{\text{change}} + \mathcal{L}_{\text{sem}} + \mathcal{L}_{\text{sa}}, & \text{for SCD}
\end{cases}
\end{equation}
where $\mathcal{L}_{\text{change}}$ is the pixel-wise cross-entropy loss on the binary change map and is used in both BCD and SCD. 
In the SCD setting, $\mathcal{L}_{\text{sem}}$ denotes the semantic segmentation loss for the two temporal masks (also implemented as pixel-wise cross-entropy loss), 
and $\mathcal{L}_{\text{sa}}$ is a semantic alignment loss based on cosine similarity.

\textbf{(ii) Bi-temporal Reconstruction Contrast.}
To align the original and reconstructed features at the token level, 
we adopt an InfoNCE-style contrastive loss within each batch.
Let $I_i,\hat{I}_i \in \mathbb{R}^{B\times L\times D_v}$ be the original and reconstructed features for temporal phase $i\!\in\!\{1,2\}$, 
where $B$ is the batch size, $L$ is the number of tokens per sample and $D_v$ is the feature dimension. 
For each token in $I_i$, its counterpart at the same spatial position in $\hat{I}_i$ is treated as the positive, 
and all other reconstructed tokens in the batch serve as negatives.  
The contrastive loss for phase $i$ is
\vspace{-2mm}
\begin{equation}
\mathcal{L}(I_i,\hat{I}_i)
= \frac{-1}{BL} \sum_{b=1}^{B}\sum_{l=1}^{L}
\log
\frac{\exp\big(\cos(I_i^{b,l},\hat{I}_i^{b,l})/\tau\big)}
{\sum_{b',l'} \exp\big(\cos(I_i^{b,l},\hat{I}_i^{b',l'})/\tau\big)}
% \vspace{-1mm}
\end{equation}
where $\cos(\cdot,\cdot)$ denotes cosine similarity and $\tau$ is a temperature hyper-parameter.  
The bi-temporal reconstruction contrast is then defined as
\vspace{-2mm}
\begin{equation}
\mathcal{L}_{\text{recon}} =
\mathcal{L}(I_1^4,\hat{I}_1^4) +
\mathcal{L}(I_2^4,\hat{I}_2^4)
\vspace{-2mm}
\end{equation}

\textbf{(iii) Transition Constraint.}
To enhance the discriminability of the transition features between changed and unchanged regions, we introduce a token-level contrastive constraint between $\Delta_1$ and $\Delta_2$.  
Let $\Delta_{1},\Delta_{2} \in \mathbb{R}^{L\times D}$ denote the transition tokens from the two temporal branches at corresponding spatial positions, and $y_l\!\in\!\{+1,-1\}$ be the token label at location $l$ indicating unchanged ($+1$) or changed ($-1$) regions.  
The constraint is formulated as
\vspace{-2mm}
\begin{equation}
\mathcal{L}_{\mathrm{trans}}
= \frac{1}{L} \sum_{l=1}^{L}
\begin{cases}
1 - \cos(\Delta_{1,l}, \Delta_{2,l}), & y_l = +1 \\[4pt]
\max\big(0, \cos(\Delta_{1,l}, \Delta_{2,l})\big), & y_l = -1
\end{cases}
% \vspace{-2mm}
\end{equation}
which encourages transition tokens at unchanged locations to be similar, while driving them apart at changed ones.

\textbf{(iv) Final Training Objective.}
The total training loss is defined as
\vspace{-2mm}
\begin{equation}
\mathcal{L}_{\mathrm{total}} =
\mathcal{L}_{\mathrm{cd}} +
\lambda_1 \mathcal{L}_{\mathrm{recon}} +
\lambda_2 \mathcal{L}_{\mathrm{trans}}
\vspace{-2mm}
\end{equation}
where $\lambda_1$ and $\lambda_2$ are coefficients that balance the two auxiliary regularization terms.
\section{Experiment}

\subsection{Experimental Setup}

% lcy：
We evaluate the proposed method on two types of change detection tasks using six publicly available datasets: SCD and BCD. For SCD, we adopt SECOND \cite{yang2021second}, JL1-SCD, and HRSCD \cite{daudt2019hrscd}; for BCD, we use LEVIR-CD \cite{chen2020levircd}, WHU-CD \cite{ji2018whucd}, and SYSU-CD \cite{shi2021sysucd}.
Specifically, SECOND (0.5–3 m resolution) comprises 4,662 image pairs across six land-cover categories; JL1-SCD (0.75 m) includes 8,000 pairs covering six categories; and HRSCD (0.5 m) is a large-scale dataset containing 291 bi-temporal image patches with six semantic classes.
For BCD, LEVIR-CD (0.5 m) serves as a standard building-change benchmark with 637 image pairs; WHU-CD (0.3 m) is typically divided into cropped patches, producing approximately 7,620 patch pairs from large aerial scenes; and SYSU-CD (0.5 m) consists of 20,000 patch pairs spanning urban and rural regions.

% xxxxxxxxxxxxxxxxxxxxxxxxxx
For both SCD and BCD tasks, we employ a unified three-branch decoder architecture comprising two semantic segmentation branches and one binary change detection branch.
In the SCD setting, all three branches are jointly supervised to learn per-temporal semantics and the corresponding change mask.
In the BCD setting, only the binary change detection branch is directly supervised, while the semantic branches serve as auxiliary feature regularizers.
All decoders adopt a UNet-style architecture.
The model is implemented in PyTorch and trained on a single NVIDIA RTX 4090 GPU. We use the Adam optimizer with an initial learning rate of $1\times 10^{-4}$, $\beta_1=0.9$, $\beta_2=0.999$, and a linear decay schedule. The batch size is set to 8, and training is conducted for 200 epochs.
% The batch size is set to 8, and training proceeds for 200 epochs.

\begin{table*}[t]
\centering
\caption{Performance comparison of different semantic change detection methods on SECOND, JL1-SCD and HRSCD datasets. \dag denotes the results with our re-implementation. The best results are \textbf{bolded} and the second-best results are \underline{underlined}. 
% All results of the three evaluation metrics are described as percentages.
}
\vspace{-3mm}
\label{tab:scd_sota}
% \footnotesize
\scriptsize
\begin{tabular}{l|ccc|ccc|ccc|cc}
\toprule
& \multicolumn{3}{c|}{\textbf{SECOND}} & \multicolumn{3}{c|}{\textbf{JL1-SCD}} & \multicolumn{3}{c|}{\textbf{HRSCD}} & & \\
Methods & \textbf{mIoU} & \textbf{SeK} & \textbf{F$_{\text{scd}}$} & \textbf{mIoU} & \textbf{SeK} & \textbf{F$_{\text{scd}}$} & \textbf{mIoU} & \textbf{SeK} & \textbf{F$_{\text{scd}}$} &\textbf{Params} &\textbf{GFLOPs} \\
\midrule
HRSCD-str4\dag$_{\text{19}}$ \cite{daudt2019hrscd} & 71.08 & 18.71 & 59.01 & 76.73 & 30.20 & 69.68 & 63.37 & 5.22 & 49.64 & 13.71 & 43.97\\
SSCDl\dag$_{\text{22}}$ \cite{ding2022bisrnet} & 73.23 & 23.03 & 63.13 & 86.54 & 57.78 & 87.01 & 67.07 & 19.13 & 67.58 & 23.31 & 189.76\\
Bi-SRNet\dag$_{\text{22}}$ \cite{ding2022bisrnet} & 73.26 & 23.14 & 63.25 & 86.63 & 58.27 & 87.28 & 66.95 & 20.17 & 68.68 &23.38 &190.30\\
TED\dag$_{\text{24}}$ \cite{ding2024scannet} & 73.27 & 23.26 & 63.21 & 86.41 & 57.21 & 86.62 & 66.68 & 18.42 & 66.97 & 24.19 & 204.29\\
SCanNet\dag$_{\text{24}}$ \cite{ding2024scannet} & 73.43 & 23.58 & 63.82 & 87.61 & 61.69 & 88.95 & 68.28 & 22.11 & 70.53 &27.90 &264.95\\
DEFO-MLTSCD\dag$_{\text{24}}$ \cite{li2024defo} & \underline{73.71} & 23.91 & 63.65 & 86.94 & 58.93 & 87.54 & 66.58 & 18.51 & 67.16 &26.02 &401.09\\
CdSCNet\dag$_{\text{24}}$ \cite{wang2024cdsc} & 73.26 & 23.53 & 63.77 & 84.18 & 52.01 & 85.78 & 68.48 & 22.01 & 70.39 &33.86 &134.80\\
ChangeMamba\dag$_{\text{24}}$ \cite{chen2024changemamba} & 73.47 & 23.93 & 63.97 & \underline{87.75} & \underline{62.58} & \underline{89.52} & \underline{68.82} & \underline{23.12} & \underline{71.44} &89.99 &146.70\\
LSAFNet\dag$_{\text{25}}$ \cite{zhou2024lsafnet} & 73.66 & \underline{24.33} & \underline{64.52} & 85.81 & 57.66 & 87.64 & 67.34 & 20.66 & 69.48 &27.86 &521.92\\
\midrule
TaCo (ours) & \textbf{73.77} & \textbf{24.73} & \textbf{64.93} & \textbf{88.13} & \textbf{63.20} & \textbf{89.61} & \textbf{69.25} & \textbf{23.71} & \textbf{71.89} &23.75 &108.57\\
\bottomrule
\end{tabular}
% \vspace{-3mm}
\end{table*}

% \vspace{-6mm}

\begin{table*}[t]
\centering
\caption{Performance comparison of different binary change detection methods on LEVIR-CD, WHU-CD and SYSU-CD datasets. \dag denotes the results with our re-implementation. The best results are \textbf{bolded} and the second-best results are \underline{underlined}. 
% All results of the three evaluation metrics are described as percentages.
}
\vspace{-3mm}
\label{tab:bcd_sota}
% \footnotesize
\scriptsize
\begin{tabular}{l|ccc|ccc|ccc|cc}
\toprule
& \multicolumn{3}{c|}{\textbf{LEVIR-CD}} & \multicolumn{3}{c|}{\textbf{WHU-CD}} &\multicolumn{3}{c|}{\textbf{SYSU-CD}} & &\\
Methods & \textbf{F$_1$} & \textbf{IoU} & \textbf{OA} & \textbf{F$_1$} & \textbf{IoU} & \textbf{OA} & \textbf{F$_1$} & \textbf{IoU} & \textbf{OA} & \textbf{Params} & \textbf{GFLOPs} \\
\midrule
IFNet\dag$_{\text{20}}$ \cite{zhang2020ifnet} & 88.11 & 78.75 & 98.81 & 89.73 & 81.37 & 99.20 &80.23 & 66.98 & 91.21 &50.71 &41.10 \\
SNUNet\dag$_{\text{21}}$ \cite{fang2021snunet} & 88.59 & 79.51 & 98.85 & 87.14 & 77.22 & 98.95 &70.59 &54.54 & 88.60 &1.35 &4.72  \\
SwinUnet\dag$_{\text{22}}$ \cite{zhang2022swinsunet} & 87.77 & 78.21 & 98.77 & 89.93 & 81.71 & 99.22& 77.78&63.64 &90.07 &30.28 &11.83  \\
BIT\dag$_{\text{22}}$ \cite{chen2021bit} & 90.03 & 81.87 & 99.01 & 91.90 & 85.01 & 99.35 &78.73 &64.93 & 89.72 &3.04 &8.75  \\
ChangeFormer\dag$_{\text{22}}$ \cite{bandara2022changeformer} & 88.83 & 79.90 & 98.88 & 90.30 & 82.32 & 99.26 & 78.01 & 63.95 & 89.53 &41.02 &202.78  \\
MSCANet\dag$_{\text{22}}$ \cite{liu2022mscanet} & 89.36 & 80.77 & 98.92 & 91.27 & 83.94 & 99.32 & 77.91 & 63.81 & 90.04 &16.42 &14.80  \\
Paformer\dag$_{\text{22}}$ \cite{liu2022paformer} & 89.68 & 81.29 & 98.96 & 92.29 & 85.69 & 99.40 &78.19 & 64.20 & 90.46 &16.13 &10.85  \\
DARNet\dag$_{\text{22}}$ \cite{li2022darnet} & 90.56 & 82.76 & 99.05 & 91.58 & 84.46 & 99.33 & 80.96 & 68.01 & 91.27 &15.09 &64.68  \\
ACABFNet\dag$_{\text{23}}$ \cite{song2022acabfnet}  & 89.18 & 80.48 & 98.91 & 91.21 & 83.84 & 99.31 & 81.29 & 68.48 & 91.42 &102.32 &28.28 \\
RS-Mamba\dag$_{\text{24}}$ \cite{zhao2024rsmamba} & 89.77 & 81.44 & 98.97 & 92.79 & 86.55 & 99.44 &77.07 & 62.69 & 89.79 &51.95 &123.79   \\
ChangeMamba\dag$_{\text{24}}$ \cite{chen2024changemamba} & 90.16 & 82.09 & 99.01 & 92.55& 86.13 & 99.42 & 81.12 & 68.24 & 91.50 &48.57 &114.82  \\
CDMamba\dag$_{\text{25}}$ \cite{zhang2025cdmamba} & 90.75 & 83.07 & 99.06 & \underline{93.76} & \underline{88.26} & \underline{99.51} & 78.63 & 64.78 & 90.31 & 11.90 &29.64  \\
CDxLSTM$_{\text{25}}$ \cite{wu2025cdxlstm} & \underline{90.89} & \underline{83.30} & \underline{99.07} & 92.85 & 86.61 & 99.44 & \underline{81.97} & \underline{69.45} & \underline{91.54}& 16.19 &3.92  \\
\midrule
TaCo (ours) & \textbf{90.95} & \textbf{83.41} & \textbf{99.10} & \textbf{94.27}& \textbf{89.16} & \textbf{99.55} & \textbf{83.92} & \textbf{72.30} & \textbf{92.56} &8.13 &14.40   \\
\bottomrule
\end{tabular}
% \vspace{-4mm}
\end{table*}

\begin{figure}%[b]
\centering
\includegraphics[width=3.2in]{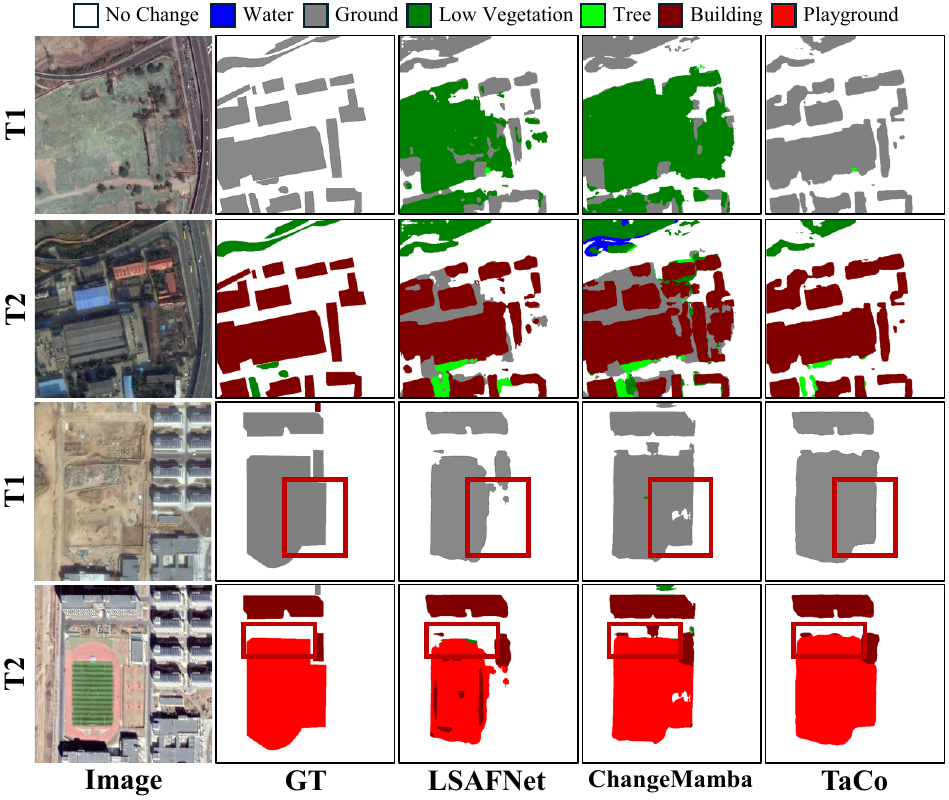}
\vspace{-3mm}
\caption{Visualization results on the SECOND dataset.}
\label{scd_vis}
\vspace{-3mm}
\end{figure}

\begin{figure}
\centering
\includegraphics[width=3.2in]{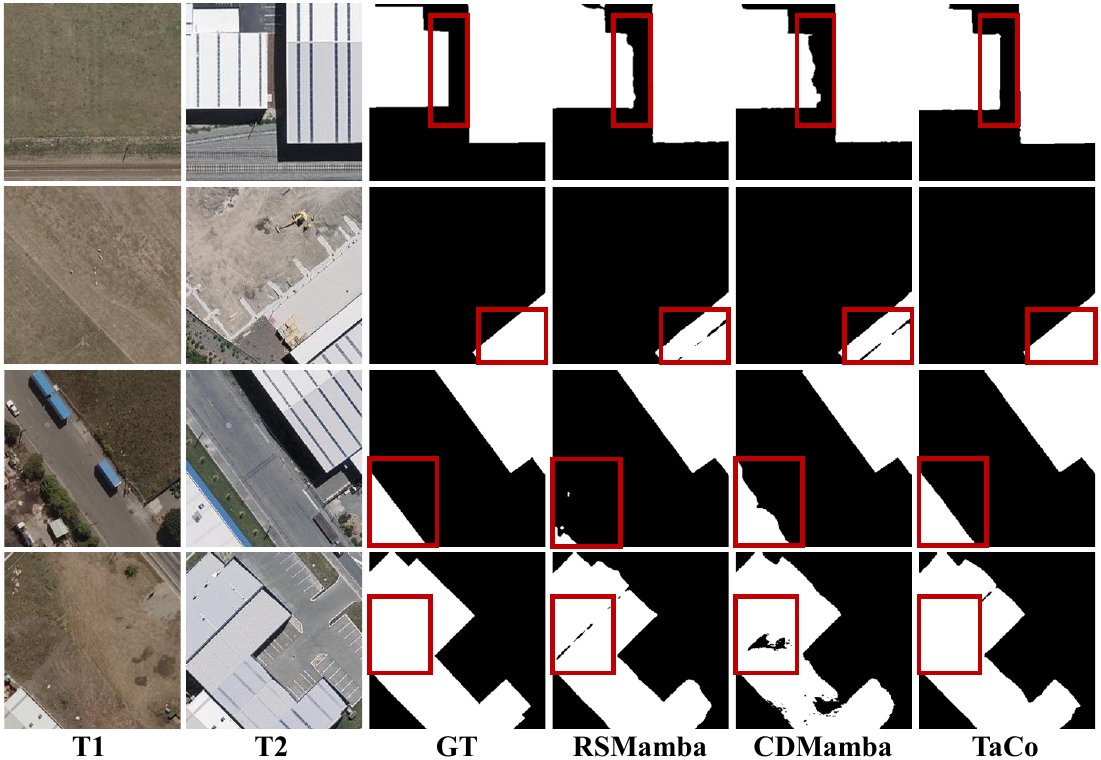}
\vspace{-2mm}
\caption{Visualization results on the WHU-CD dataset.}
\label{bcd_vis}
\vspace{-7mm}
\end{figure}

\subsection{Comparison with State-of-the-Art Methods}
We compare our approach with some representative methods across the two CD tasks. Specifically, for SCD, we include HRSCD-str4, SSCDl, Bi-SRNet, TED, SCanNet, DEFO-MLTSCD, CdSCNet, ChangeMamba, and LSAFNet. For BCD, we evaluate against IFNet, SNUNet, SwinUnet, BIT, ChangeFormer, MSCANet, Paformer, DARNet, ACABFNet, RS-Mamba, ChangeMamba, CDMamba and CDxLSTM.
As shown in Tables \ref{tab:scd_sota}–\ref{tab:bcd_sota}, TaCo achieves SOTA performance across all six datasets spanning the two CD tasks, while remaining computationally efficient during inference. This improvement stems from the proposed spatio-temporal semantic joint constraint, which enhances the model's performance without introducing additional computational overhead during inference.

Fig.~\ref{scd_vis} and \ref{bcd_vis} present qualitative results. TaCo produces cleaner and more coherent change maps with sharper boundaries and fewer false alarms.
Compared with other methods, TaCo effectively distinguishes fine-grained land-cover categories and preserves the structural integrity of large-scale changes.
Benefiting from the proposed spatio-temporal semantic joint constraint, TaCo better models temporal semantic transitions and captures semantic changes while maintaining spatial coherence in complex scenes.
These qualitative results confirm that our method enhances both semantic discrimination and temporal consistency, leading to clearer change representations.

\begin{table}%[t]
\footnotesize
\centering
\caption{Performance comparison of different losses on the SECOND and WHU-CD datasets. Here, we set $\lambda_1$= 0.1 and $\lambda_2$= 1.}
\vspace{-3mm}
\label{tab:loss_config}
\renewcommand{\arraystretch}{1.05}
\begin{tabular}{lcc}
\toprule
\textbf{Loss Configuration} & \textbf{SECOND (SeK)} & \textbf{WHU-CD (F$_1$)} \\
\midrule
$\mathcal{L}_{\mathrm{cd}}$ & 23.77 & 93.41 \\
$\mathcal{L}_{\mathrm{cd}} +
\lambda_1 \mathcal{L}_{\mathrm{recon}}$ & 24.35 & 93.85 \\
$\mathcal{L}_{\mathrm{cd}} +
\lambda_1 \mathcal{L}_{\mathrm{recon}} + \lambda_2
\mathcal{L}_{\mathrm{trans}}$ & \textbf{24.54} & \textbf{94.02} \\
\bottomrule
\end{tabular}
\vspace{-4mm}
% \vspace{2mm}
% {\scriptsize
% $\lambda_1 \mathcal{L}_{\mathrm{recon}}=0.1\times\mathcal{L}_{\text{cd}}$,\quad $\lambda_2 =1.$
% }
% \vspace{2mm}
% \raggedright{\footnotesize
% Here, 
% $\lambda_1 \mathcal{L}_{\mathrm{recon}}=0.1\times\mathcal{L}_{\text{cd}}$ and $\lambda_2 =1.$}
\end{table}

\vspace{-1mm}
\subsection{Ablation Study}
% \vspace{-1mm}
\textbf{Bi-Temporal Reconstruction Contrast and Transition Constraint.}
% Table~\ref{tab:loss_config} reports an ablation on the loss design of our spatio-temporal semantic modeling. 
% Using only the spatial detection loss $\mathcal{L}_{\text{cd}}$ denotes the baseline. 
% Adding the bi-temporal reconstruction contrast loss $\mathcal{L}_{\text{recon}}$ consistently improves SeK and F$_1$, indicating that encouraging agreement between original and reconstructed features stabilizes the temporal representations. 
% When the transition loss $\mathcal{L}_{\text{trans}}$ is further incorporated, the performance is further improved, showing that the two token-level objectives are complementary: $\mathcal{L}_{\text{recon}}$ regularizes the reconstruction of each phase, whereas $\mathcal{L}_{\text{trans}}$ enhances the semantic separability between changed and unchanged tokens. By default, we set $\lambda_1 = 0.1$ and $\lambda_2 = 1$ in all experiments.
Table~\ref{tab:loss_config} presents an ablation study on the loss components used in our spatio-temporal semantic modeling.
Using only the spatial change detection loss $\mathcal{L}_{\mathrm{cd}}$ serves as the baseline.
Introducing the bi-temporal reconstruction contrast loss $\mathcal{L}_{\mathrm{recon}}$ consistently improves both SeK and F$_1$ metrics, demonstrating that enforcing agreement between original and reconstructed features helps enhance bi-temporal representations.
Further adding the transition loss $\mathcal{L}_{\mathrm{trans}}$ yields additional gains, indicating that the proposed two token-level constraints are complementary. $\mathcal{L}_{\mathrm{recon}}$ regularizes the fidelity of reconstructed features for each temporal phase, while $\mathcal{L}_{\mathrm{trans}}$ promotes semantic discrimination between changed and unchanged tokens. 
% Unless otherwise specified, 

\begin{figure}
\centering
\includegraphics[width=3.2in]{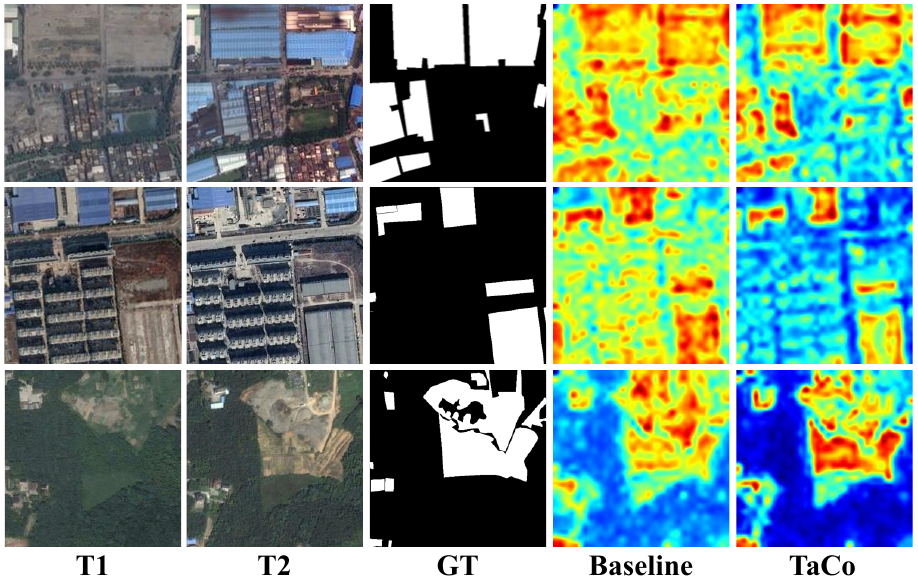}
\vspace{-3mm}
\caption{Visualization results of the differential feature maps on the SECOND dataset.}
\label{feature_vis}
\vspace{-4mm}
\end{figure}

\begin{table}
\footnotesize
% \small
\centering
\caption{Ablation study of Adaptive Semantic Integration on the SECOND and WHU-CD datasets.
``Params" refers to the number of activated parameters during training, not during inference.}
\vspace{-2mm}
\label{tab:adapter}
\begin{tabular}{l|cc|cc}
\toprule
\textbf{Methods} & \textbf{SeK ↑\tiny (SECOND)} & \textbf{Params} & \textbf{F$_1$ ↑\tiny (WHU-CD)} & \textbf{Params} \\
\midrule
w/o. ASI & 24.54 & 30.64M & 94.02 & 14.89M \\
w/. ASI & \textbf{24.73} & 31.04M & \textbf{94.27} & 15.29M \\
\bottomrule
\end{tabular}
\vspace{-5mm}
\end{table}

% \textbf{Effectiveness of the Soft Mixture-of-Expert.}
% Table~\ref{tab:adapter} evaluates the effect of the proposed SoftMoE on both accuracy and model size. 
% Compared with the backbone baseline, introducing text-guided modules already brings clear gains on SECOND and WHU-CD. 
% When the Text-guided Transition Generator is used without the adapter (\emph{w/o Adapter}), performance is improved but the alignment between textual and visual features is only implicitly learned. 
% Adding the Text Adapter (\emph{TaCo}) further boosts SeK and F$_1$, while the number of trainable parameters increases only slightly (about +0.4M). 
% This shows that the adapter offers more effective text–visual alignment and semantic modulation at a very modest parameter overhead.
\textbf{Effectiveness of Adaptive Semantic Integration.}
Table~\ref{tab:adapter} evaluates the impact of the proposed Adaptive Semantic Integration.
Using ASI yields consistent gains on both datasets, improving SeK from 24.54 to 24.73 on SECOND and F$_1$ from 94.02 to 94.27 on WHU-CD, while the number of trainable parameters increases by only about 0.4M.
These results indicate that ASI provides more semantic modulation at a very modest training parameter overhead.

\textbf{Qualitative Analysis of Spatio-Temporal Semantic Modeling}.
To further illustrate the impact of spatio-temporal semantic modeling, we visualize the enhanced stage-4 differential feature maps in the CD decoder. As shown in Fig.~\ref{feature_vis}, compared with the baseline, TaCo exhibits better activations in true change regions while effectively suppressing responses in irrelevant background areas. These observations suggest that the proposed modeling approach guides the network to focus on semantically meaningful structural changes, resulting in clearer change localization and reduced noise in unchanged regions.

\begin{figure}
\centering
\includegraphics[width=3.2in]{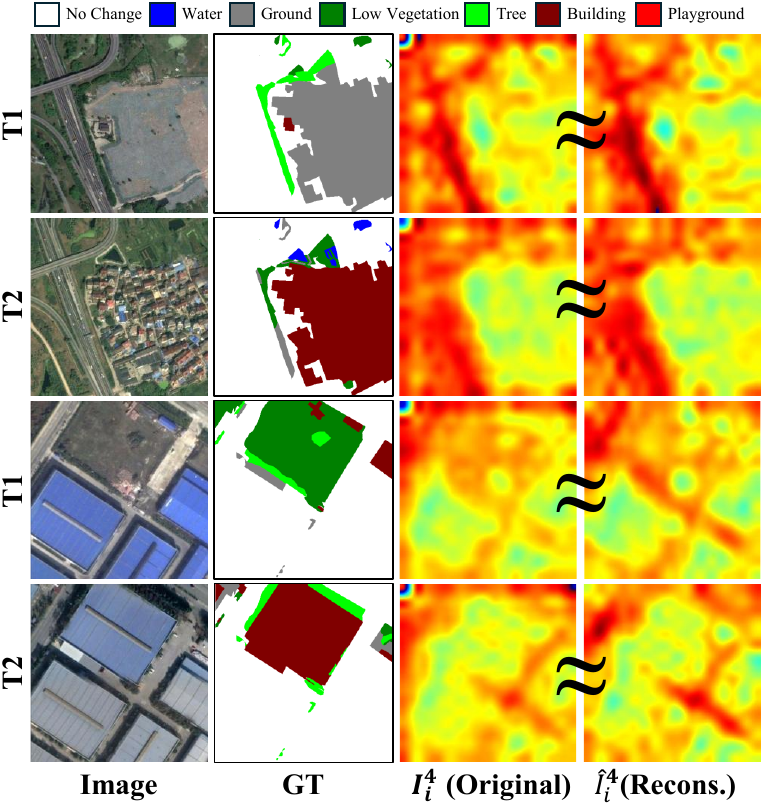}
\vspace{-3mm}
\caption{Visualization comparison of $I_i^4$ and reconstructed $\hat{I}_i^4$ under unified visualization setting.}
\label{sim_vis}
\vspace{-4mm}
\end{figure}

In Fig.~\ref{sim_vis}, we further visualize the original high-level features $I_i^4$ and their reconstructed counterparts $\hat{I}_i^4$ produced under the spatio-temporal semantic joint constraint.
The two maps exhibit strong visual correspondence across most regions. $\hat{I}_i^4$ preserves the structural layout and dominant semantic patterns of $I_i^4$, indicating that the reconstruction mechanism can reliably regenerate one temporal feature map from the other when guided by the learned transition features. This demonstrates the effectiveness of our constraint design in regulating the evolution of bi-temporal features, and confirms that the missing semantic information between temporal states is largely recoverable. 
% Our constraint regulates the evolution of bi-temporal features and enhances the discriminability of change-related representations.

\vspace{-1mm}
\subsection{Deeper into Spatio-Temporal Constraint.}
\textbf{Directionality of Spatio-Temporal Semantic Constraint Configuration.}
We first examine how the directionality (Eq.~\ref{equ_1} or Eq.~\ref{equ_2}) of the spatio-temporal semantic constraint influences model performance. 
Table~\ref{tab:text} reports results on SECOND and WHU-CD under four settings: a baseline model without any spatio-temporal semantic constraint, two one-way variants that impose a single directional transition constraint (Eq.~\ref{equ_1} or Eq.~\ref{equ_2}), and a two-way configuration that combines both directions. 
Compared with the baseline, both one-way settings consistently improve performance on SECOND and WHU-CD, showing that incorporating temporal transition modeling between bi-temporal features helps capture inter-temporal relationships beyond what can be learned from spatial mask supervision alone.

\begin{table}%[t]
\centering
\footnotesize
\caption{Directionality analysis of constraint on SECOND and WHU-CD datasets.}
\vspace{-3mm}
\label{tab:text}
\begin{tabular}{l c c}
\toprule
\textbf{Methods} & \textbf{SECOND (SeK ↑)} & \textbf{WHU-CD (F$_1$ ↑)} \\
\midrule
Baseline & 23.83 & 93.41 \\
One-way (Eq.~\ref{equ_1}) & 24.01 & 93.99 \\
One-way (Eq.~\ref{equ_2}) & 24.06 & \textbf{94.10} \\
\midrule
Two-way (Eq.~\ref{equ_1}, \ref{equ_2}) & \textbf{24.35} & 93.85 \\
\bottomrule
\end{tabular}
\vspace{-4mm}
\end{table}

Under the full two-way configuration, the model achieves the best SeK score on SECOND but does not yield the highest F$_1$ on WHU-CD. 
This difference reflects the distinct nature of the two tasks. 
SECOND is a semantic change detection benchmark with bi-temporal land-cover categories, where bidirectional constraints provide richer supervision by capturing mutual temporal transitions between categories. 
In contrast, WHU-CD is a binary building change detection dataset, where most changes correspond to a single semantic transition (e.g., building construction or demolition) and are supervised by binary masks. 
In this setting, a one-way constraint is generally sufficient to capture the relevant temporal information, and enforcing bidirectional regularization brings little additional benefit.

These observations suggest that the choice of constraint directionality should be adapted to the underlying task. 
For SCD, where categories undergo diverse and often reversible transitions, using a two-way constraint is more favorable as it provides more complete temporal supervision. 
For BCD that inherently lacks semantic translation, a single-direction constraint may be a simpler and more stable design.

\textbf{Sensitivity Analysis to Change Region Size.}
We observe that the performance gain brought by our spatio-temporal semantic modeling varies noticeably across datasets. The proposed spatio-temporal constraint strategy yields clear improvements on SECOND and WHU-CD (as shown in Table~\ref{tab:loss_config}), yet brings only marginal gains on LEVIR-CD (as shown in Table~\ref{tab:change_scale}). We attribute this discrepancy to differences in the relative size and visual prominence of changed regions. SECOND and WHU-CD contain large-area changes in high-resolution imagery. In contrast, changes in LEVIR-CD occupy only a small portion of each $256 \times 256$ patch, making them substantially less visually salient. To validate this hypothesis, we conduct two complementary analyses on binary (LEVIR-CD) and semantic (SECOND) change detection tasks.

On LEVIR-CD, we train the model using cropped images ($128 \times 128$) rather than ($256 \times 256$) so that changed buildings occupy a larger fraction of each image. This manipulation directly increases the visual salience of change regions. As reported in Table~\ref{tab:change_scale}, the performance improvement brought by our spatio-temporal semantic modeling becomes significantly larger under the $128 \times 128$ setting (e.g., +0.13 F$_1$), compared with the negligible gain observed in the original $256 \times 256$ setup (only +0.03 F$_1$). This result provides direct evidence that the effectiveness of our constraint is strongly correlated with the relative proportion of changed areas.

We further analyze SECOND dataset by grouping samples according to their per-class change ratios. As shown in Table~\ref{tab:oa_large_small}, TaCo yields substantially larger OA improvements on samples with large or spatially extensive change regions (e.g., +2.80 for Building, +3.48 for Low Vegetation), whereas gains on small change regions are more limited.
This again confirms that the proposed constraint benefits most when the changed structures are sufficiently large.

% The size sensitivity can be explained from the perspective of feature formation and constraint design.
% Our spatio-temporal semantic modeling operates on low-resolution, high-level feature maps and is guided by class-level textual prompts. When the true change occupies only a very small portion of the feature map:

The size sensitivity can be explained from the perspective of feature formation and constraint design.
Our spatio-temporal semantic modeling operates on low-resolution, high-level feature maps and is guided by class-level textual semantics. When the true change occupies only a very small portion of the feature map:
(1) Transition features become difficult to infer.
Small changes may be overwhelmed by the surrounding context, increasing the difficulty for the TTG module to synthesize accurate transition representations.
(2) Loss contributions of small regions are diluted.
In the joint constraint, the reconstruction and transition losses assigned to small change areas constitute only a minor proportion of the overall objective, weakening their influence during optimization.
When local changes are limited, the network tends to focus on globally stable semantic cues rather than subtle structural transitions, reducing the utility of the proposed temporal semantic constraints.

In summary, our analysis demonstrates that the effectiveness of the spatio-temporal semantic modeling is positively correlated with the scale and prominence of the changed regions. In future work, we will explore more refined mechanisms to substantially improve the performance and robustness of the model on small change instances as well.

\begin{table}%[t]
\centering
\caption{Sensitivity Analysis to the relative size of change regions on LEVIR-CD dataset. }
\vspace{-2mm}
\label{tab:change_scale}
\footnotesize
\begin{tabular}{l|cc}
\toprule
\textbf{Training Setting} & \textbf{F$_1$} & \textbf{IoU} \\
\midrule
Baseline (256$\times$256) & 90.92 & 83.36 \\
+ Constraint Strategy (256$\times$256) & 90.95 & 83.41 \\
Improvement $\Delta$ & 0.03 & 0.05 \\
\midrule
Random Crop (128$\times$128) & 91.64 & 84.56 \\
+ Constraint Strategy (128$\times$128) & 91.77 & 84.79 \\
Improvement $\Delta$ & \textbf{0.13} & \textbf{0.23} \\
\bottomrule
\end{tabular}
% \vspace{-1mm}
\end{table}

\begin{table}%[t]
\centering
\caption{
% Performance comparison between baseline and TaCo on SECOND dataset under different change ratio and semantic categories. 
% Metrics are macro-averaged overall accuracies (OA) per image. “Large” and “Small” denote samples with class area ratio $>$25\% and $<$5\%, respectively.
Sensitivity Analysis to the relative change size on the SECOND dataset for different change-ratio semantic categories.
}
\vspace{-2mm}
\label{tab:oa_large_small}
\footnotesize
\setlength{\tabcolsep}{4pt}
\begin{tabular}{l|ccc|ccc}
\toprule
\multirow{2}{*}{\textbf{Category}} &
\multicolumn{3}{c|}{\textbf{Small ($<$5\%)}} &
\multicolumn{3}{c}{\textbf{Large ($>$25\%)}} \\
& \textbf{Base} & \textbf{TaCo} & $\Delta$ & \textbf{Base} & \textbf{TaCo} & $\Delta$ \\
\midrule
Building & 53.81 & 55.15 & +1.35 & 68.36 & 71.16 & \textbf{+2.80} \\
Low Vegetation & 32.08 & 34.71 & +2.63 & 57.76 & 61.24 & \textbf{+3.48}\\
\bottomrule
\end{tabular}
\vspace{-6mm}
\end{table}

% \section{Conclusion}
% In this work, we presented TaCo, which brings spatio-temporal semantic modeling into RSCD. 
% Built upon a mask-supervised siamese encoder–decoder architecture, TaCo augments the baseline with a Text-guided Transition Generator and a Spatio-Temporal Semantic Joint Constraint, thereby modeling change as a semantic transition process rather than a purely spatial difference driven only by mask supervision.
% Extensive experiments on six public benchmarks, covering both binary and semantic change detection, show that TaCo consistently outperforms state-of-the-art methods, while maintaining modest parameter overhead and competitive inference efficiency. 
% Beyond numerical gains, TaCo offers a systematic way to incorporate high-level textual priors into temporal feature modeling, leading to more stable spatio-temporal representations of change.
% In future work, we plan to extend TaCo toward a unified paradigm for multi-modal and multi-temporal change understanding, with a focus on improving robustness and scalability in real-world monitoring scenarios.

% \vspace{-1mm}
\section{Conclusion}
% \vspace{-1mm}
Unlike previous methods that primarily focus on spatial differences under a mask-supervised framework, we propose TaCo, which treats change as a semantic transition between bi-temporal states and explicitly models spatio-temporal semantic consistency by leveraging learned transition features from textual prompts. The proposed consistency strategy can effectively improve the model’s performance without introducing any additional computational overhead during inference.
Extensive experiments on six public benchmarks, covering binary and semantic change detection, demonstrate that TaCo consistently outperforms previous SOTA methods. In addition, we provide an in-depth analysis of the proposed spatio-temporal consistency constraints, offering valuable insights for future research.

% In future work, we plan to further develop TaCo into a more general framework for multi-modal and multi-temporal change detection.

% To address this limitation, we propose TaCo, a spatio-temporal semantic consistent network, which extends the existing mask-supervised framework with a spatio-temporal semantic joint constraint.
% TaCo views changes as a semantic transition, where one temporal state can be derived from another through transition features.
% Specifically, we design a Text–guided Transition Generator that integrates textual semantics with bi-temporal visual features to construct the cross-temporal transition features.
% Besides, we propose a spatio-temporal semantic joint constraint, composed of bi-temporal reconstruct constraints and a transition constraint, where the former aligns reconstructed and original features, and the latter enhances discrimination for changes. This strategy can effectively improve the model’s performance without introducing any additional computational overhead during inference.
% Extensive experiments on six public datasets, spanning both binary and semantic change detection tasks, demonstrate that TaCo achieves SOTA performance.
% \clearpage
% {
%     \small
%     \bibliographystyle{ieeenat_fullname}
%     \bibliography{main}
% }
\clearpage
\setcounter{page}{1}
\maketitlesupplementary

\section*{A. Detailed Architecture of TaCo}

TaCo follows a siamese encoder–decoder architecture that processes bi-temporal images $(X_1, X_2)$ through a shared backbone and two lightweight decoder branches. 

The two temporal images are first encoded by a weight-shared SegFormer~\cite{xie2021segformer} backbone, 
which produces multi-level features ${I_i^1, I_i^2, I_i^3, I_i^4}$ for temporal phase $i$. 
Among them, the stage-4 representation $I_i^4$ serves as the input to the spatio-temporal semantic modeling.

% TaCo adopts a siamese encoder architecture where bi-temporal images $(X_1, X_2)$ are processed by a weight-shared SegFormer~\cite{xie2021segformer} backbone. 
% Each encoder produces multi-level features ${I_i^1, I_i^2, I_i^3, I_i^4}$, with stage-4 feature $I_i^4$ serving as the input to the Spatio-Temporal Semantic Modeling.

\textbf{Seg Decoders.}
For each temporal phase (see Fig.~\ref{overview}(a) in the main paper), we fuse the stage-3 and stage-4 features by upsampling the deeper feature and concatenating it with the shallower one, followed by a $1{\times}1$ convolution to unify channel dimensions.
A top–down decoder then progressively upsamples this fused feature and combines it with lower-stage features (via upsampling and concatenation), producing a set of refined multi-scale representations $I_i^1$, $I_i^2$, and $I_i^4$ for temporal phase $i$.
In the SCD setting, $I_i^1$ is fed into a lightweight prediction head to generate the Seg mask for each temporal image.

\textbf{CD Decoder.}
As illustrated in Fig.~\ref{overview}(a) of the main paper, the CD decoder operates on the refined features from the two semantic decoders.
It first constructs a deep change representation by concatenating the refined high-level features $I_1^4$, $I_2^4$, and their absolute difference $|I_1^4 - I_2^4|$, following~\cite{li2024defo}.
This deep change feature is then progressively upsampled and fused with absolute differences from lower-level refined features (e.g., $|I_1^2 - I_2^2|$, $|I_1^1 - I_2^1|$).
The final fused representation is processed by a shallow prediction head to produce the CD mask.

\section*{B. Dataset Details}

SECOND~\cite{yang2021second} is a well-annotated semantic change detection benchmark constructed from high-resolution satellite images over several Chinese cities, including Hangzhou, Chengdu and Shanghai. Each bi-temporal pair is provided as a $512\times512$ RGB patch with spatial resolution ranging from 0.5\,m to 3\,m. The dataset defines six land-cover categories (tree, building, playground, low vegetation, ground and water) and provides dense semantic change labels for both temporal phases. In our experiments, we follow the official split using 2,968 image pairs for training and 1,694 pairs for testing.

JL1-SCD is a semantic change detection dataset tailored to cropland monitoring, collected from the JiLin-1 remote sensing satellite. It consists of $256\times256$ image pairs with a spatial resolution better than 0.75 m.
The dataset defines five land-cover classes—cropland, non-vegetated ground surface (n.v.g. surface), road, building, and forest/grassland—providing sufficient semantic granularity for agricultural change analysis. 
Consistent with \cite{wang2024cdsc}, we use 4{,}000 image pairs for training and 2{,}000 pairs for testing.

The HRSCD~\cite{daudt2019hrscd} dataset is a large-scale semantic change detection benchmark. It contains 291 pairs of $10{,}000\times10{,}000$ RGB aerial images acquired between 2006 and 2012, covering diverse urban and rural regions. Five land-cover classes are annotated, including artificial surfaces, agricultural areas, forests, wetlands and water. For computational efficiency, we crop the original large images into $256\times256$ patches and automatically discard patches with a negligible change ratio (less than 5\%). The remaining patches are randomly divided into training and testing subsets with an 8:2 ratio.

LEVIR-CD~\cite{chen2020levircd} is a binary building change detection dataset constructed from multi-temporal Google Earth imagery. It comprises 637 bi-temporal pairs with a spatial resolution of 0.5~m and a patch size of $1024\times1024$, with acquisition intervals ranging from 5--14~years. The dataset focuses on man-made changes, and each pair is annotated with pixel-level building change masks. In our pipeline, the original pairs are further divided into $256\times256$ patches, and we adopt a split of 7{,}120/1{,}024/2{,}048 patches for training, validation and testing, respectively.

WHU-CD~\cite{ji2018whucd} is another building-oriented binary change detection benchmark based on very high-resolution aerial imagery. It is collected over the Christchurch area in New Zealand, with images acquired in 2012 and 2016, and provides pixel-wise labels indicating where buildings have appeared or disappeared. During preprocessing, we crop the images into $256\times256$ patches and randomly split them into 6{,}096/762/762 patches for training, validation and testing.

SYSU-CD~\cite{shi2021sysucd} is a large-scale binary change detection dataset comprising 20{,}000 pairs of $256\times256$ aerial images with 0.5~m spatial resolution acquired over Hong Kong between 2007 and 2014. The dominant changes correspond to typical urban development patterns, including newly built urban buildings, suburban dilation, groundwork before construction, vegetation changes, road expansion and sea construction. We follow the commonly used split and adopt 8{,}000/4{,}000/8{,}000 image pairs for training, validation, and testing, respectively.

\section*{C. Implementation Details}

All components in our spatio-temporal semantic modeling are configured with lightweight and unified hyper-parameters.
Within the Text-guided Transition Generator, Adaptive Semantic Integration uses $M{=}6$ experts to adapt the textual category priors to a unified text–vision fusion space of dimension $D{=}256$.
The subsequent cross-modal fusion is performed using a transformer decoder with $L{=}6$ layers and 4 attention heads, which provides sufficient capacity for injecting textual semantics into temporal features while keeping computation modest.
For the bi-temporal reconstruction constraints, we adopt a fixed temperature $\tau{=}0.07$, which offers stable optimization across all datasets.

\begin{figure*}[t]
\centering
\includegraphics[width=0.85\linewidth]{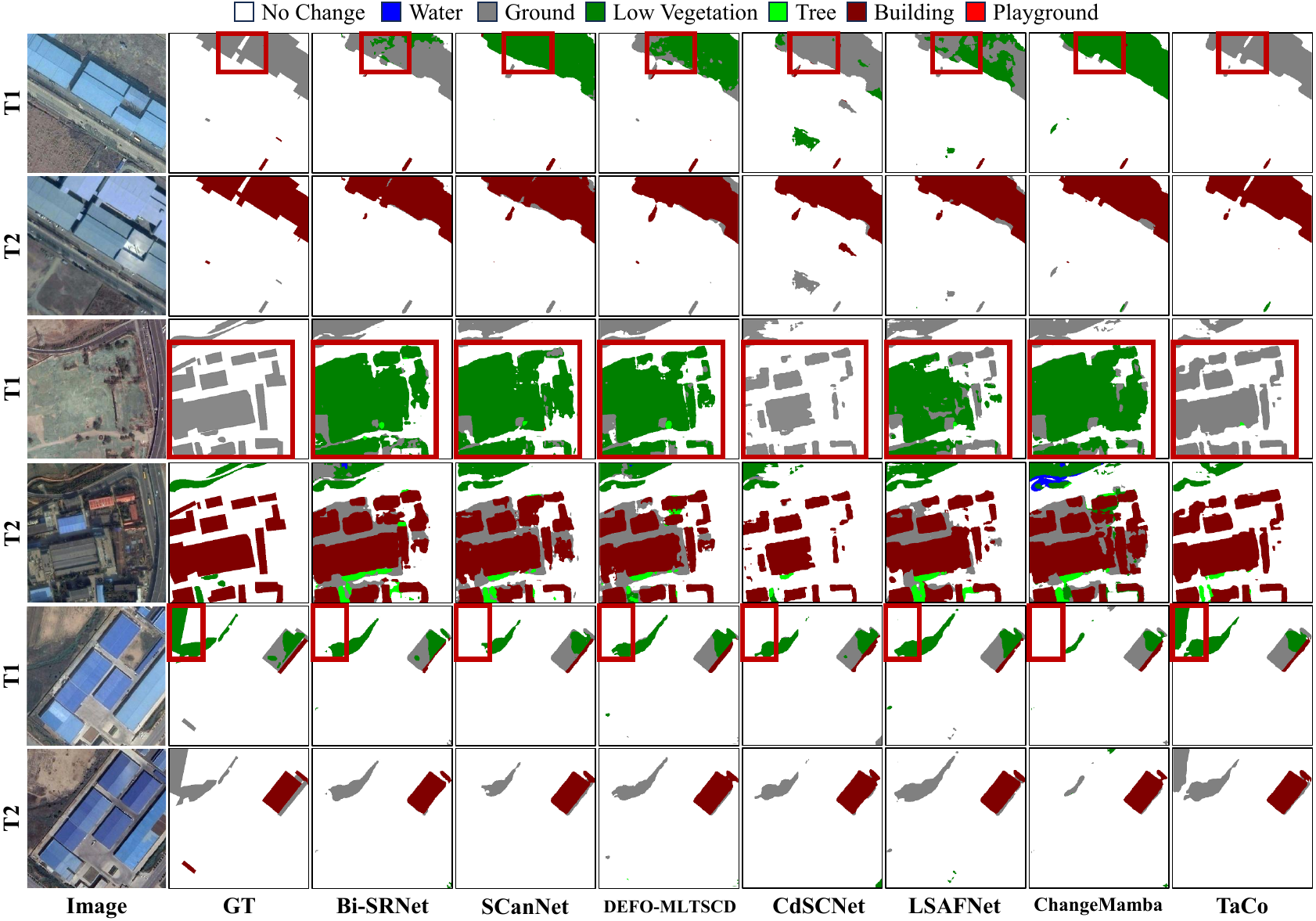}
\vspace{-2mm}
\caption{Visualization results on the SECOND dataset.}
\label{second}
\vspace{-2mm}
\end{figure*}

\begin{figure*}
\centering
\includegraphics[width=0.85\linewidth]{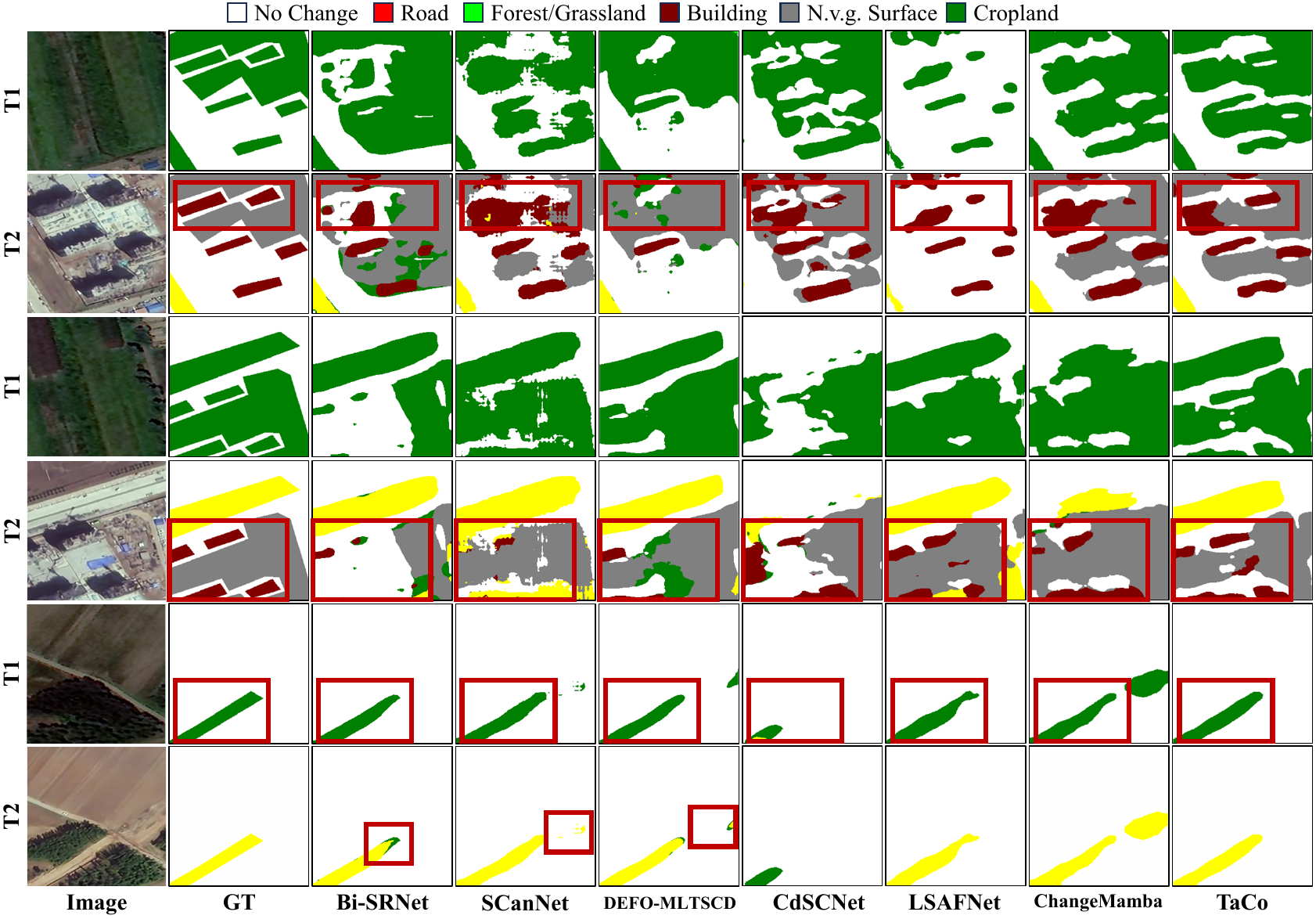}
\vspace{-2mm}
\caption{Visualization results on the JL1-SCD dataset.}
\label{jl1}
% \vspace{-6mm}
\end{figure*}

\begin{figure*}
\centering
\includegraphics[width=0.85\linewidth]{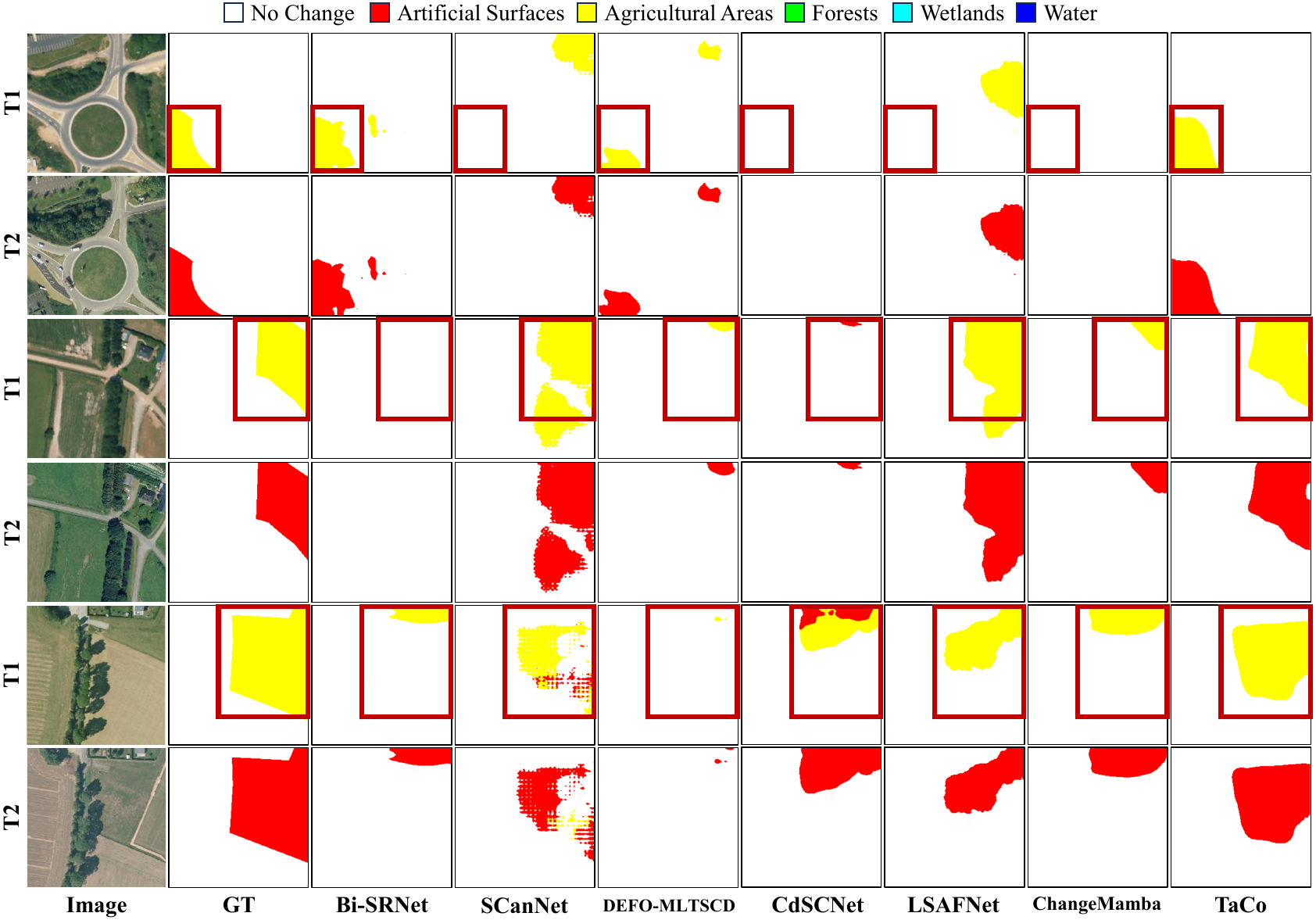}
\vspace{-2mm}
\caption{Visualization results on the HRSCD dataset.}
\label{hrscd}
\vspace{-2mm}
\end{figure*}

\begin{figure*}
\centering
\includegraphics[width=0.85\linewidth]{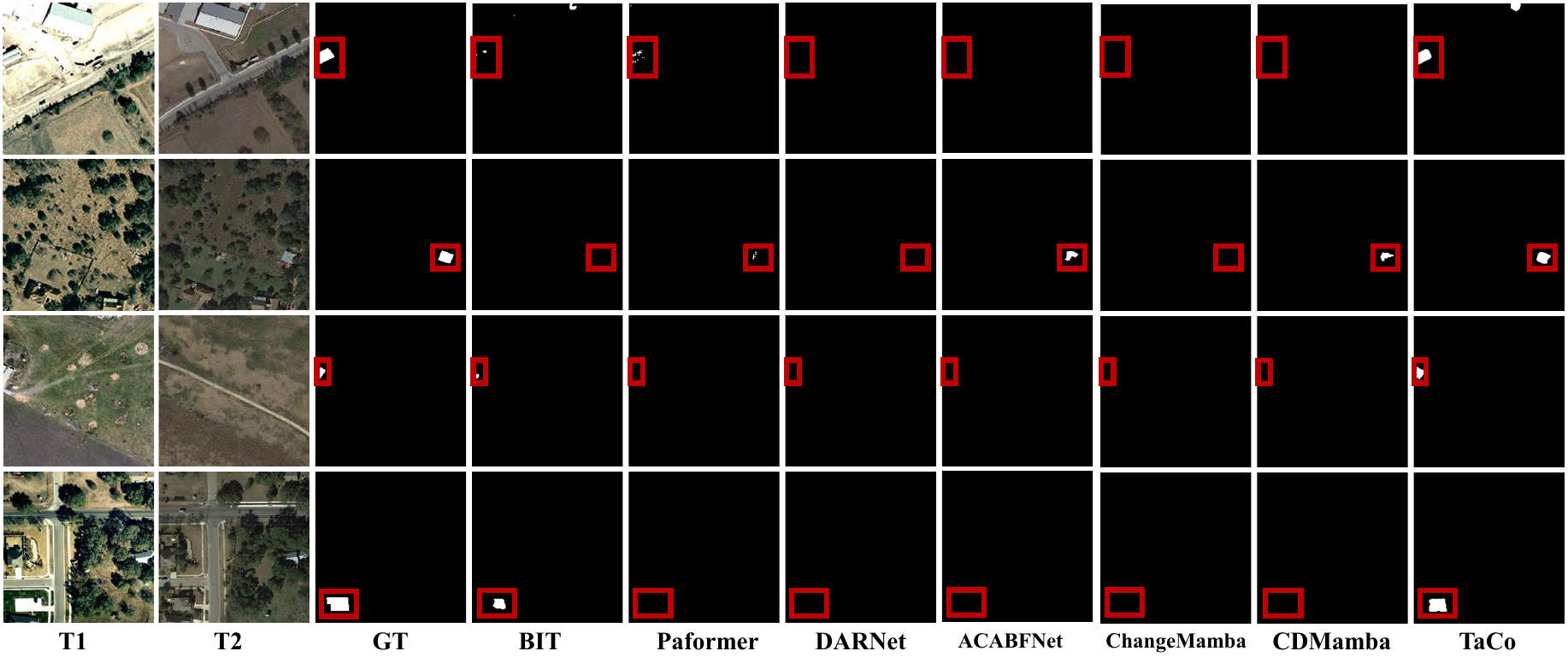}
\vspace{-2mm}
\caption{Visualization results on the LEVIR-CD dataset.}
\label{levir}
\vspace{-2mm}
\end{figure*}

\begin{figure*}
\centering
\includegraphics[width=0.85\linewidth]{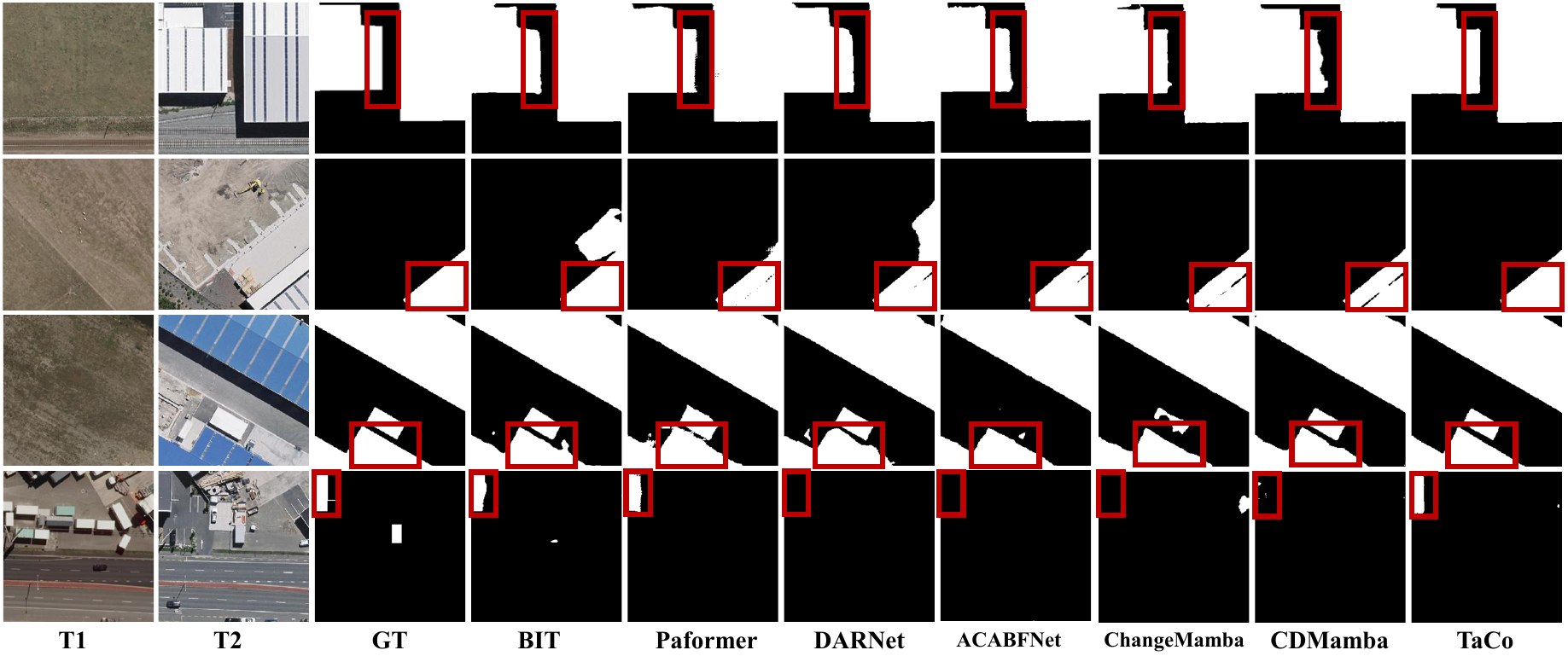}
\vspace{-2mm}
\caption{Visualization results on the WHU-CD dataset.}
\label{whu}
\vspace{-2mm}
\end{figure*}

\begin{figure*}
\centering
\includegraphics[width=0.85\linewidth]{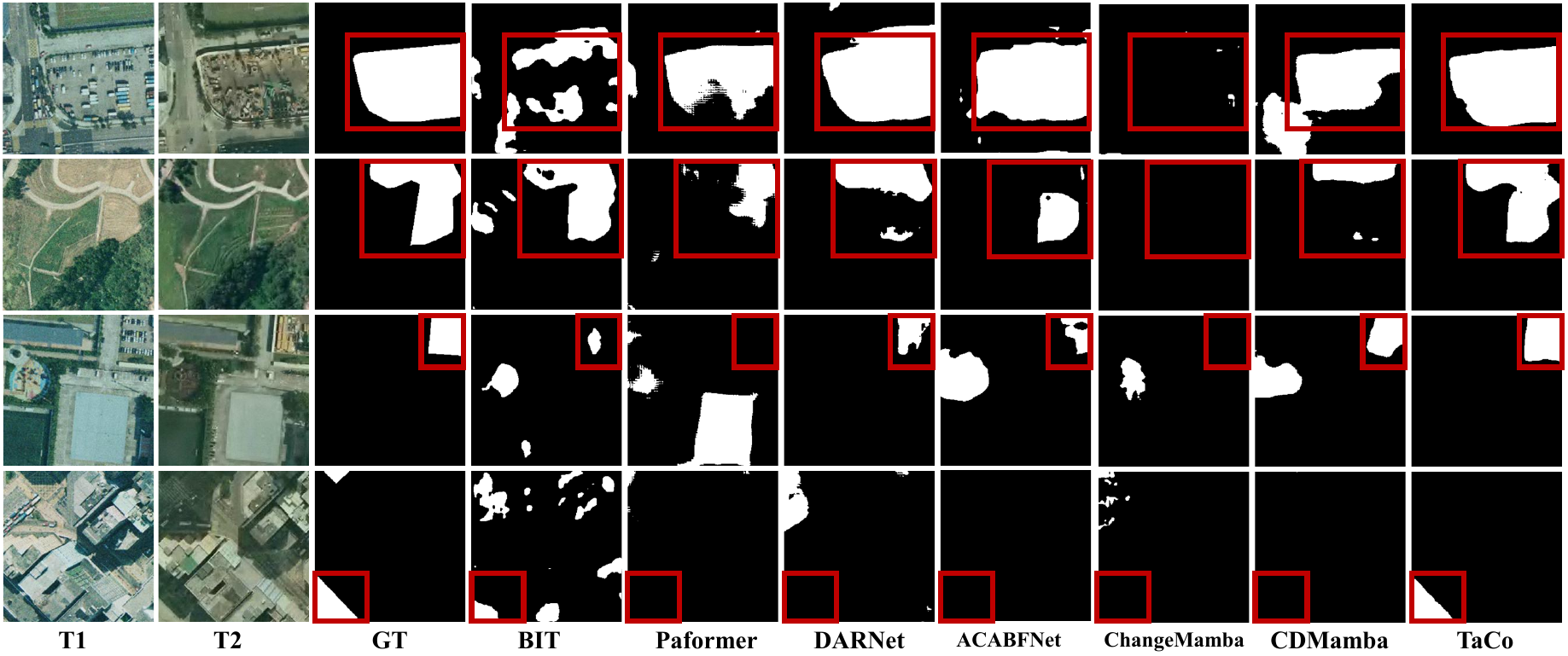}
\vspace{-2mm}
\caption{Visualization results on the SYSU-CD dataset.}
\label{sysu}
\vspace{-2mm}
\end{figure*}

\section*{D. Additional Visualization Results}

We provide extensive qualitative comparisons across all six datasets, as shown in Fig.~\ref{second}-\ref{sysu}.  
These visualizations demonstrate that TaCo effectively refines unchanged regions and enhances semantic differentiation in changed areas, validating the usefulness of spatio-temporal semantic modeling.

\clearpage
{
    \small
    \bibliographystyle{ieeenat_fullname}
    \bibliography{main}
}


\end{document}